\documentclass[10pt,sigconf]{acmart}
\makeatletter
\def\@ACM@checkaffil{
    \if@ACM@instpresent\else
    \ClassWarningNoLine{\@classname}{No institution present for an affiliation}%
    \fi
    \if@ACM@citypresent\else
    \ClassWarningNoLine{\@classname}{No city present for an affiliation}%
    \fi
    \if@ACM@countrypresent\else
        \ClassWarningNoLine{\@classname}{No country present for an affiliation}%
    \fi
}
\makeatother

\acmConference[GNNet '23]{Proceedings of the 2nd Graph Neural Networking Workshop 2023}{December 8, 2023}{Paris, France}

\AtBeginDocument{%
  \providecommand\BibTeX{{%
    \normalfont B\kern-0.5em{\scshape i\kern-0.25em b}\kern-0.8em\TeX}}}

\usepackage{amsmath,amsfonts}
\usepackage{graphicx}
\usepackage{caption}  
\usepackage{subcaption}  
\usepackage{textcomp}
\usepackage{color, colortbl}  
\usepackage{algorithm}  
\usepackage[noend]{algpseudocode}  
\usepackage{enumitem}
\usepackage{balance}
\usepackage{comment}

\renewcommand\footnotetextcopyrightpermission[1]{} 
\setcopyright{none}
\newcommand\blfootnote[1]{%
  \begingroup
  \renewcommand\thefootnote{}\footnote{#1}%
  \addtocounter{footnote}{-1}%
  \endgroup
}

\graphicspath{ {./images/} }
\begin{document}
\title[Detecting Contextual Network Anomalies with GNNs]{Detecting Contextual Network Anomalies\\with Graph Neural Networks}

\author{Hamid Latif}
\email{hamid.latif@upc.edu}
\affiliation{%
  \institution{Barcelona Neural Networking Center,\\Universitat Politècnica de Catalunya, Spain}
}
\author{José Suárez-Varela}
\email{jose.suarez-varela@telefonica.com}
\affiliation{%
  \institution{Telefónica Research, Spain}
}
\author{Albert Cabellos-Aparicio}
\email{alberto.cabellos@upc.edu}
\affiliation{%
  \institution{Barcelona Neural Networking Center,\\Universitat Politècnica de Catalunya, Spain}
}
\author{Pere Barlet-Ros}
\email{pere.barlet@upc.edu}
\affiliation{%
  \institution{Barcelona Neural Networking Center,\\Universitat Politècnica de Catalunya, Spain}
}

\renewcommand{\shortauthors}{Hamid Latif, José Suárez-Varela, Albert Cabellos-Aparicio, \& Pere Barlet-Ros}

\begin{abstract}

Detecting anomalies on network traffic is a complex task due to the massive amount of traffic flows in today's networks, as well as the highly-dynamic nature of traffic over time. In this paper, we propose the use of Graph Neural Networks (GNN) for network traffic anomaly detection. We formulate the problem as \textit{contextual anomaly detection} on network traffic measurements, and propose a custom GNN-based solution that detects traffic anomalies on origin-destination flows. In our evaluation, we use real-world data from Abilene (6~months), and make a comparison with other widely used methods for the same task (PCA, EWMA, RNN). The results show that the anomalies detected by our solution are quite complementary to those captured by the baselines (with a max. of 36.33\% overlapping anomalies for PCA). Moreover, we manually inspect the anomalies detected by our method, and find that a large portion of them can be visually validated by a network expert (64\% with high confidence, 18\% with mid confidence, 18\% normal traffic). Lastly, we analyze the characteristics of the anomalies through two paradigmatic cases that are quite representative of the bulk of anomalies.

\end{abstract}

\maketitle

\section{Introduction}
\blfootnote{If you cite this paper, please use the CoNEXT '23 reference: Hamid Latif, José Suárez-Varela, Albert Cabellos-Aparicio, and Pere Barlet-Ros. 2023. Detecting Contextual Network Anomalies with Graph Neural Networks. In Proceedings of the 2nd Graph Neural Networking Workshop 2023 (GNNet ’23). Association for Computing Machinery, New York, NY, USA. \url{https://doi.org/10.1145/3630049.3630171}}

Anomaly detection is a crucial task for network operators. It enables the identification of unusual events, some of them caused by equipment failures, sub-optimal configurations, cyber attacks, or sudden changes in the traffic workload~\cite{lakhina2004diagnosing}. As such, it permits the early detection and troubleshooting of network issues that otherwise might remain unnoticed until they have severe impact on the end-user experience.

In this paper, we tackle the problem of network traffic anomaly detection in backbone networks (e.g., ISPs, NRENs). This problem has been largely studied in the past~\cite{ahmed2016survey}, and it is typically formulated as a data-driven use case where the objective is the detection of anomalies on network traffic measurements. A common approach is to rely on traffic matrices with updated measurements at the Origin-Destination (OD) flow level~\cite{zhang2005network}. Typically, anomaly detection tools employ statistical models to determine the \textit{normal} behavior of traffic flows: then they apply these models to find deviations with respect to the learned distributions~\cite{lakhina2004diagnosing,gonzalez2022dc}.

Detecting network anomalies in the wild still remains a complex task, due to the vast amount of traffic flows generated and terminated over time, and the noisy and highly-dynamic nature of network traffic~\cite{bhuyan2013network}. More importantly, alarms generated by anomaly detection tools are usually followed by a manual inspection from network experts, which requires a thorough analysis to determine if the anomaly pinpoints a real network issue. This makes it imperative to limit the volume of false alarms generated by these systems, as it translates to direct OPEX savings for the operator.

In general, data anomalies can be divided into different classes \mbox{---~e.g.,} point, contextual, collective \cite{bhuyan2013network}\mbox{~---} all of them considering different criteria to define how the \textit{normality} model is built. In this paper, we focus on \textit{contextual anomaly detection} methods for time series, which may help detect different types of anomalies often unnoticed by other methods~\cite{dimopoulos2017detecting}. Particularly, we consider contextual anomalies over OD flows, that is, we consider that an anomaly exists if the traffic of a flow starts to deviate from other flows with similar past activity (i.e., the \textit{context flows}). Despite the interesting properties of contextual methods, defining the context between elements is not trivial, as it may depend on different criteria, such as the distance metric used. More recent works propose the use of Machine Learning techniques to learn the context in an unsupervised manner~\cite{gonzalez2022dc}.

In this paper, we propose the use of Graph Neural Networks (GNN)~\cite{battaglia2018relational} for network traffic anomaly detection. We use graphs to represent OD flows as sets of elements interconnected. Then we use a Graph Attention-based model~\cite{velivckovic2017graph} to process the traffic time series of OD flows, and extract a context of these flows based on their past traffic activity. With this information our solution can accurately capture contextual anomalies in OD flows in an unsupervised manner, by detecting traffic deviations of specific flows from their respective context flows.

To evaluate our solution we use real-world data from Abilene, covering a period of six months~\cite{zhang2005network}. We train our solution on the first three months, and then test it over the next three months of data, obtaining a total of 600 anomalies generated by our method (50 per week \cite{zhang2005network}). To assess the practicality of our solution, we compare the previous anomalies with those detected by other well-known network anomaly detection methods: Principal Component Analysis (PCA)~\cite{lakhina2004diagnosing}, Exponentially Weighted Moving Average (EWMA)~\cite{tang2014adaptive} and Recurrent Neural Networks (RNN)~\cite{goh2017anomaly}. All these methods are tested on the same data, and configured to produce 600 anomalies. The results show that at least 63\% of the anomalies of our method are not individually detected by the baselines. This suggests that our solution can be a good complement to the other non-contextual detection methods. To further assess the correctness of the anomalies detected by our solution, we take a set with 100 randomly picked anomalies, and expose them to a network expert to visually annotate them. The results show that a total of 64\%, 18\%, and 18\% are respectively labeled as high-confidence anomaly, mid-confidence anomaly, and normal traffic. Lastly, we delve into two example cases that illustrate how our method amplifies the detection of contextual anomalies, thus achieving a considerably higher detection rate in this type of anomalies when we compare it to the other baselines.

\section{Related Work}

Network traffic anomaly detection has been widely studied in the past~\cite{ahmed2016survey}, with some popular works such as~\cite{lakhina2004diagnosing, zhang2005network}, that apply PCA to detect anomalies on network-wide traffic.

In the literature we can also find some works for contextual anomaly detection on time series data. These methods first look for sets of elements that have similar past activity (i.e., the contexts); then they define anomalies as deviations of a target element with respect to its context. For example, \cite{dimopoulos2017detecting} tackles the problem of detecting performance anomalies on end-users in broadband networks, and \cite{hayes2014contextual} combines content and contextual detection methods to find anomalies in sensor monitoring data.

More recent approaches use Deep Learning (DL) to deal with the complexity behind the vast amount of data collected from networks. For example, in~\cite{gonzalez2022dc} they propose a model combining dilated convolutional neural networks and a variational autoencoder, dubbed DC-VAE, and apply it to anomaly detection on multivariate time series from mobile networks.

GNNs are a subclass of DL models specifically designed to work on graph-structured data~\cite{battaglia2018relational}, and have been successfully applied to many different fields, including chemistry, biology, or computer vision~\cite{zhou2020graph}. In the field of communication networks, GNNs have been applied to a variety of tasks~\cite{suarez2022graph}, such as routing optimization~\cite{rusek2019unveiling}, power control in wireless networks~\cite{shen2020graph}, Distributed Denial of Service (DDoS) attack detection~\cite{li2022graphddos}, or network intrusion detection~\cite{pujol2022unveiling}.

Some recent works propose to apply these models to contextual anomaly detection. For example, in \cite{hu2020graph} the authors use a GNN-based autoencoder for anomaly detection in real-world mobility datasets. Likewise, in \cite{deng2021graph} they perform contextual anomaly detection on multivariate time series from a distributed sensor network, and apply it to find anomalies in water treatment and distribution plants.

In \cite{xu2021anomaly} the authors propose using Transformers to perform time series anomaly detection. Contrary to our solution, they focus on exploiting temporal correlations, both local (i.e., prior-association) and global (i.e., series-association). Likewise, the authors of \cite{tuli2022tranad} propose the use of Transformers for multivariate time series anomaly detection.

\section{Our solution}

\begin{figure}[!t]
    \centerline{\includegraphics[scale=0.3]{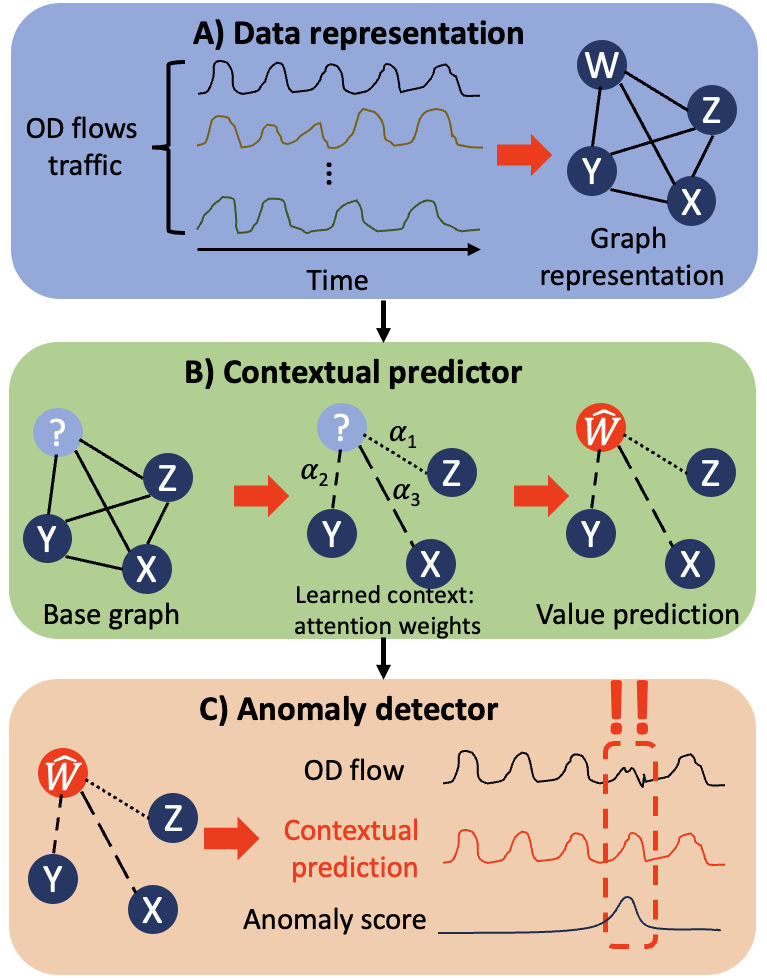}}
    \caption{Overview of our solution.}
    \label{Overview}
\end{figure}

Fig. \ref{Overview} shows an overview of the proposed solution. It is divided into three main blocks: 
\begin{itemize}[leftmargin=*]
    \item \textbf{Data representation}: Converts traffic measurement data into graph-structured representations, where graph nodes contain the OD flow traffic measurements during a specific time period $T$.
    \item \textbf{Contextual predictor}: Learns the context of a target OD flow and uses it to predict the next value using only information from the context flows at $t$+$1$.
    \item \textbf{Anomaly detector}: Applies a detection method to classify anomalous deviations in OD flows with respect to their context.
\end{itemize}

\subsection{Data representation}\label{sec:data-representation}

Our method is designed to process OD flow traffic measurements as input~\cite{zhang2005network}, which are represented by matrix $\mathbb{R}^{MxT}$, where $M$ is the number of OD flows in the network, and $T$ is the number of temporal samples per flow. We create a graph representation $G$ for every time sample $t$ (see Fig.~\ref{Overview}-A). We represent each OD flow as a graph node $f_i$, and connect flows with graph edges $e_{ij}$ that represent their relationships (i.e., the context). Since the context of flows is initially unknown, we connect flows all-to-all; then we use a contextual predictor that determines the flows' context.

\subsection{Contextual predictor}\label{sec:contextual-predictor}

\setlength{\textfloatsep}{0.4cm}
\begin{algorithm}[!t]
\footnotesize{
\caption{Contextual predictor}
\label{Contextual predictor pseudocode}
\begin{flushleft}
\hspace*{\algorithmicindent} \textbf{Input:} $F, P, O$\\
\end{flushleft}
\begin{algorithmic}[1]
\State \textbf{for each} $f \in F$ \textbf{do} $h_f^0 \gets MLP_F(f)$
\State \textbf{for each} $p \in P$ \textbf{do} $h_p^0 \gets MLP_P(p)$
\State \textbf{for each} $o \in O$ \textbf{do} $h_o^0 \gets MLP_O(o)$
\For {$p_i \in P,\ p_j \in N(i),\ o_i \in O,\ o_j \in N(i)$}
    \State $\alpha_{i,j} \gets Attention([h_{p_i}, h_{o_i}], [h_{p_j}, h_{o_j}])$ 
\EndFor
\For {$k=0$ to $(K-1)$}
    \For {$f \in F$}
        \State ${m_f}^{k+1} \gets \sum_{f_{j} \in N(i)} \hspace*{1mm} h_{f_j}^k \times \alpha_{ij}$
        \State ${h_f}^{k+1} \gets U(h_f^k, m_f^{k+1})$
    \EndFor
\EndFor
\State $\hat{f^{t+1}} \gets R_f(h_f^K)$
\end{algorithmic}
}
\end{algorithm}

A main feature of our method is that it learns the underlying flow contexts. To do so, we use a GNN model that learns to set attention weights $\alpha_{i,j}$ on the edges between flows (see Fig.~\ref{Overview}-B). These edges represent the relevance of the connected flows to their context. Our GNN model comprises three main inputs: $(i)$ the flow traffic measurements in a given time window $[t-W, t]$, $(ii)$ a positional encoding that uniquely identifies flows (one-hot encoding), and $(iii)$ a flow label that identifies the OD flow where the anomaly detection will be done (one-hot encoding).\\

Three main ideas motivate the design of our contextual predictor:
\begin{itemize}[leftmargin=*]
    \item \textbf{Positional encoding}: We need to identify uniquely flows to then define in the model which ones are related between them, and what is the relevance to their context.
    \item \textbf{Target flow label}: We need the model to identify the flow where the contextual prediction is made. Then, it predicts the next value in the flow ($t$+$1$) by considering only measurements from its context flows at $t$+$1$.
    \item \textbf{Attention weights}: Each flow is related to the rest of flows of its context, but different flows may have different degrees of relation. For this reason, our model implements an attention mechanism (Graph Attention Layer~\cite{velivckovic2017graph}) that quantifies the level of relation between flows.
\end{itemize}

Our GNN-based model (Algorithm~\ref{Contextual predictor pseudocode}) implements a message passing process~\cite{suarez2022graph} that combines the information of flow measurements~$F$, the positional encodings $P$, and the target flow label~$O$ to predict the traffic of the target flow $\hat{f}$ at $t$+$1$. First, the hidden states of flow measurements $h_f^0$, positional encodings $h_p^0$, and target flow label $h_o^0$ are initialized (lines 1-3), by using a Multi-Layer Perceptron (MLP). Then, the hidden states of positional encodings $h_p^0$, and the target flow label $h_o^0$ are passed though the $Attention$ layer to generate attention coefficients $\alpha_{ij}$ for each flow pair (lines 4-5). Here, $N(i)$ is a function that returns the set of neighboring flows for flow $i$. Afterward, a message passing process is executed for $K$ iterations (lines 6-9). First, the model generates flow-level aggregate messages $m_f$ by combining the neighbors' flow measurement hidden states $h_f$ with the corresponding attention coefficients $\alpha_{ij}$ (line~8). Then, flow hidden states $h_f$ are updated along with their aggregated messages $m_f$ by using the $U$ function (line 9), which is implemented by a neural network. After the message passing process, the final hidden state of the target flow $h_f^K$ is passed through the Readout function $R_f$, which produces a traffic estimate for the next time step $\hat{f}^{t+1}$ (line 10).

We train the model in a self-supervised manner, using as loss function the Mean Absolute Error (MAE) between the predicted flow measurement and its actual value at $t$+$1$. This enables to train the $Attention$ function end-to-end. After training, we store the flow-level MAEs ($MAE_{tr}^{f}$), which are then used by the anomaly detector.

\subsection{Anomaly detector}\label{sec:anomaly-detector}

\begin{algorithm}[!t]
\footnotesize{
\caption{Anomaly detection score}
\label{Anomaly detector pseudocode}
\begin{flushleft}
\hspace*{\algorithmicindent} \textbf{Input:} $F, \hat{F}$ \\
\end{flushleft}
\begin{algorithmic}[1]
\For {$f \in F,\ \hat{f} \in \hat{F}$}
    \State $Err_{pred} \gets |f_i^{t+1} - \hat{f}_i^{t+1}|$
    \State $MAE_{tr} \gets MAE_{tr}^{f}$
    \State $score_f \gets \frac{Err_{pred}}{MAE_{tr} \times~\delta}$
\EndFor
\State Return $[score_{0}, ..., score_{M}]$
\end{algorithmic}
}
\end{algorithm}

Lastly, the anomaly detector generates an anomaly score at the OD flow level (see Fig.~\ref{Overview}-C). Algorithm \ref{Anomaly detector pseudocode} shows the anomaly scoring method, which is based on computing differences between the traffic estimates of the contextual predictor $\hat{F}$ and the real traffic values measured at $t$+$1$. We use a configurable detection threshold $\delta$ to tune the volume of anomalies generated. For each flow, we compute the prediction error $Err_{pred}$ (line 2). Then, we pick the MAE of the contextual predictor for flow $f$ (i.e., $MAE_{tr}^{f}$), which was stored during training (Sec.~\ref{sec:contextual-predictor}). The $MAE_{tr}^{f}$ values are used as normalization parameters to calibrate the detection sensitivity across OD flows. Finally, we produce flow-level anomaly scores by combining the prediction error $Err_{pred}$, the detection threshold $\delta$, and the calibration value $MAE_{tr}^{f}$ (line 4). All samples with an anomaly score > 1 are then considered anomalous.

\section{Evaluation}\label{Evaluation}
In this section we evaluate the proposed solution with real-world traffic measurements from the Abilene backbone network (dataset description in Sec.~\ref{Dataset}). First, we compare our solution with other well established methods for network anomaly detection (Sec.~\ref{comparison-baselines}). Then, we perform a manual analysis of the anomalies detected by our solution to assess their correctness, as well as to characterize them (Sec.~\ref{sec:analysis-anomalies}).

\subsection{Dataset and Training}\label{Dataset}

In our experiments, we use six months of traffic measurement data from the Internet2's Abilene backbone network (12 nodes, 15 links, 132 OD flows)~\cite{zhang2005network}. Measurements are collected at the OD flow granularity and sampled in 5-minute intervals. In total, the dataset contains 48,096 temporal samples per OD flow. We use the first three months of data for training and validation, and the next three months for testing (i.e., anomaly detection).

To prepare our method, we train a contextual predictor \mbox{---~i.e.,} a GNN \mbox{model~---} per OD flow (see Sec.~\ref{sec:contextual-predictor}), where each time sample uses the graph data representation in Sec.~\ref{sec:data-representation}. The GNN model comprises a Graph Attention layer \cite{velivckovic2017graph} for the $Attention$ function, and the hidden state of flows are represented as 128-element vectors. The $Update$ function is implemented with a Gated Recurrent Unit (GRU), and the Readout function $R_f$ with a 3-layer MLP with 128 units per layer and LeakyReLU activation functions. After some data analysis, we observed that traffic measurements follow a heavy-tailed distribution. Then, we apply a logarithmic normalization to reduce the variance in input/output features and make the training more stable. Lastly, we set the training length to 50 epochs of 24,048 samples each. 

\subsection{Comparison with other methods}\label{comparison-baselines}

In this section we compare our solution with other methods commonly used for network anomaly detection. 
Since the Abilene dataset does not include annotated anomalies, we follow an unsupervised approach: we analyze the overlap between the anomalies generated by the different methods considered in our evaluation. With this, we seek to gain some insight on how complementary are the sets of anomalies generated by each of them. Later, in Sec.~\ref{sec:analysis-anomalies} we manually analyze a random subset of these anomalies.

\subsubsection{\textbf{Baselines}} \hfil

We consider the following four network anomaly detection methods as a benchmark for our evaluation:

\begin{enumerate}[leftmargin=*]
    \item \textbf{PCA Links}: We apply the PCA method proposed in~\cite{lakhina2004diagnosing}, which uses link traffic measurements to detect network anomalies. This PCA method classifies network-wide anomalies at specific time bins, so we set all OD flows as anomalous at the detected time samples~\cite{ringberg2007sensitivity}.
    \item \textbf{PCA Flows}: The previous PCA method operates on link measurements, while the anomaly detection method we propose operates at the OD flow granularity. Hence, we re-implement the previous method~\cite{lakhina2004diagnosing} to detect anomalies from OD flow measurements. This new method is a direct adaptation that considers OD flow traffic as input.
    \item \textbf{EWMA}: We implement an EWMA~\cite{tang2014adaptive} to detect temporal traffic anomalies on OD flows. EWMA is a widely used method for anomaly detection on time series. The input of this model is a time series with past traffic measurements of a given OD flow. This baseline detects anomalies as deviations from its predictions, using the same criteria as in our solution to score anomalies (see Sec.~\ref{sec:anomaly-detector}).
    \item \textbf{RNN}: We implement a RNN model~\cite{bengio1994learning}, which represents a widely used DL method for anomaly detection on time series. Similar to our method, it is trained in a self-supervised manner ---~i.e., we train it to predict the next value of OD flow traffic time series. After training, we use the same criteria for scoring anomalies (Sec.~\ref{sec:anomaly-detector}). 
\end{enumerate}

\subsubsection{\textbf{Overlap analysis}} \hfill

We run all the anomaly detection methods on the test dataset (last three months). The objective of this experiment is to measure the overlap of the anomalies produced by all the solutions. This should help us better understand how complementary are the anomalies of our contextual anomaly detection method with respect to the other baselines. 

For a fair comparison, we set the detection threshold of all methods to produce the 600 most relevant anomalies (50 anomalies/week, as proposed in~\cite{zhang2005network}). After some fine tuning, we select a window of 5 time samples for the time series-based baselines (EWMA, RNN). In the dataset we found a significant amount of OD flows where the traffic volume is negligible. We consider that these flows are not of interest to our anomaly detection study, as they exhibit a very static behavior. Hence, in our evaluation we take only flows with an average traffic higher than 3 bps. Also, we group detected anomalies that are at a distance of less than 30 minutes, and consider them as only one that extends over the entire period. A main aspect to consider in our solution is that it relies on a contextual predictor (Sec.~\ref{sec:contextual-predictor}). In some cases, an OD flow may lack other flows with similar behavior, thus making it complex to find a robust context to produce accurate predictions. To discard these cases, we apply our solution to OD flows where the Mean Relative Error of the predictor in training is <30\%.
\begin{table}[!t]
\caption{Percentage of overlapping detections among all evaluation baselines (each with top 600 anomalies).}
\label{600 detections comparison}
\resizebox{\columnwidth}{!}{ %
\begin{tabular}{c c c c c c }
\toprule
                              & PCA Links & PCA Flows & EWMA    & RNN     & Ours    \\
                              \midrule
                              \midrule
PCA Links                                & -         & 37.67\%   & 18.33\% & 21\%    & 28.67\% \\ \midrule
PCA Flows                                & 57.33\%   & -         & 16\%    & 20.50\% & 36.16\% \\ \midrule
EWMA                                     & 49.17\%   & 24\%      & -       & 84.83\% & 1.67\%  \\ \midrule
RNN                                      & 49.67\%   & 28.50\%   & 71\%    & -       & 2.33\%  \\ \midrule
\rowcolor{gray!15}
Ours & 34.33\%   & 36.33\%   & 1.33\%  & 0.67\%  & -   \\
\bottomrule
\end{tabular}
}
\end{table}

Table~\ref{600 detections comparison} shows the detection overlaps between all the anomaly detection methods. Each row refers to the set of anomalies detected by a given method, while columns are the detections made by the other methods on the different anomaly sets. Looking at the sets of anomalies generated by the baselines (first four rows), we can differentiate two main groups: PCA methods have a larger overlap between them, while time series methods (EWMA and RNN) exhibit a significantly high overlap (>71\% between them). This was expected due to the similar nature of the methods in both groups. More importantly, we observe that, from the set of anomalies generated by our method (last row), the \textit{PCA Flows} baseline is the one capturing a higher portion of anomalies (overlap of~36.33\%). Overall, from these results we observe that at least 63\% of the anomalies of our solution are not individually captured by any of the baselines. This suggests that our solution produces a quite complementary set of anomalies compared to the other methods.

To further delve into these results, we test if the baselines are capable of increasing their overlap with other methods by gradually reducing their detection threshold. However, this also comes at the risk of producing a higher portion of false positives.

In these experiments we observe that the time series-based methods (EWMA and RNN) do not increase significantly their recall, reaching an overlap of 13.5\% and 7\% respectively when they produce 6x more anomalies than our solution (i.e., 3,600 anomalies). For the PCA-based methods, we observe that the volume of false alarms increases quickly to unrealistic levels, marking >20\% of the total samples as anomalous after small threshold modifications. These results expose the known sensitivity issues of PCA-based methods applied to network anomaly detection~\cite{ringberg2007sensitivity}.

\subsection{Analysis of anomalies}\label{sec:analysis-anomalies}

\begin{table}[!t]
\caption{Results of the visual validation. Results are calculated over 100 randomly selected anomalies.}
\label{manual validation}
\resizebox{0.85\columnwidth}{!}{ %
\begin{tabular}{c c c}
\toprule
                         \textbf{High confidence}     & \textbf{Mid confidence} & \textbf{Normal traffic}  \\
                              \midrule
                              \midrule
                        64 & 18 & 18 \\ 
\bottomrule
\end{tabular}
}
\end{table}

In the previous experiments we observe that our solution produces sets of anomalies complementary to those of the baselines. In this section we aim to get a better understanding of the correctness and nature of the detected anomalies.

First, we assess the relevance of the anomalies produced. Particularly, we set our evaluation to visually validating if there was an actual anomaly in the samples detected by our solution. For this validation, we randomly pick 100 anomalies out of the 600 ones generated, and expose them to a network expert to assess them. The expert is tasked to annotate anomalies according to the description of contextual anomalies described earlier in this paper ---~i.e., there is an anomaly if the behavior of the OD flow deviates with respect to its context flows. For each anomaly, we provide both the evolution of the target OD flow and the traffic of the top-5 context flows extracted from our contextual predictor. Given the complexity of performing a binary classification (i.e., normal/anomalous), the evaluation follows a softer approach that divides annotations into three tiers: $(i)$ high-confidence anomaly, $(ii)$ mid-confidence anomaly, and $(iii)$ normal traffic. Table~\ref{manual validation} shows the results of the validation. From these figures, we conjecture that the large portion of anomalies annotated with high (64\%) and mid (18\%) confidence is a good indicator that our method is actually detecting a great share of relevant network anomalies. A more comprehensive validation should be done to get more definite results on its effectiveness by using labeled real-world datasets.

We analyze in more detail two paradigmatic cases that represent a large portion of the anomalies detected by our solution. Particularly, we identify two main classes of anomalies: $(i)$ those where an OD flow deviates from its context flows, and $(ii)$ those where an OD flow remains fairly stable over time and the context flows start to diverge.

\begin{figure}[!t]
    \centerline{\includegraphics[scale=0.24]{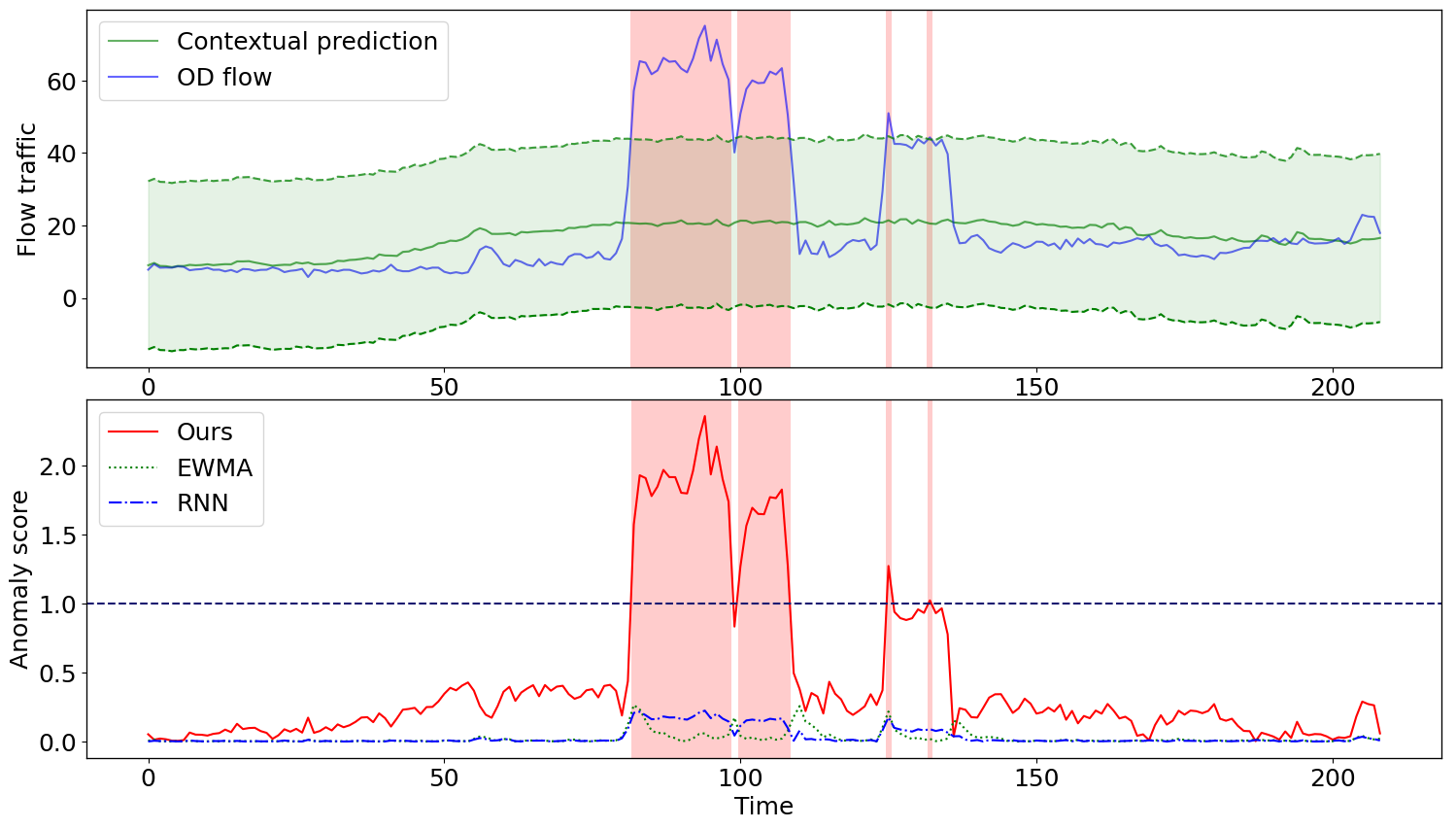}}
    \caption{Anomaly \#1 - Deviation of an OD flow w.r.t. its context flows.}
    \label{Anomaly 1 plots}
\end{figure}

\begin{figure}[!t]
    \centerline{\includegraphics[scale=0.24]{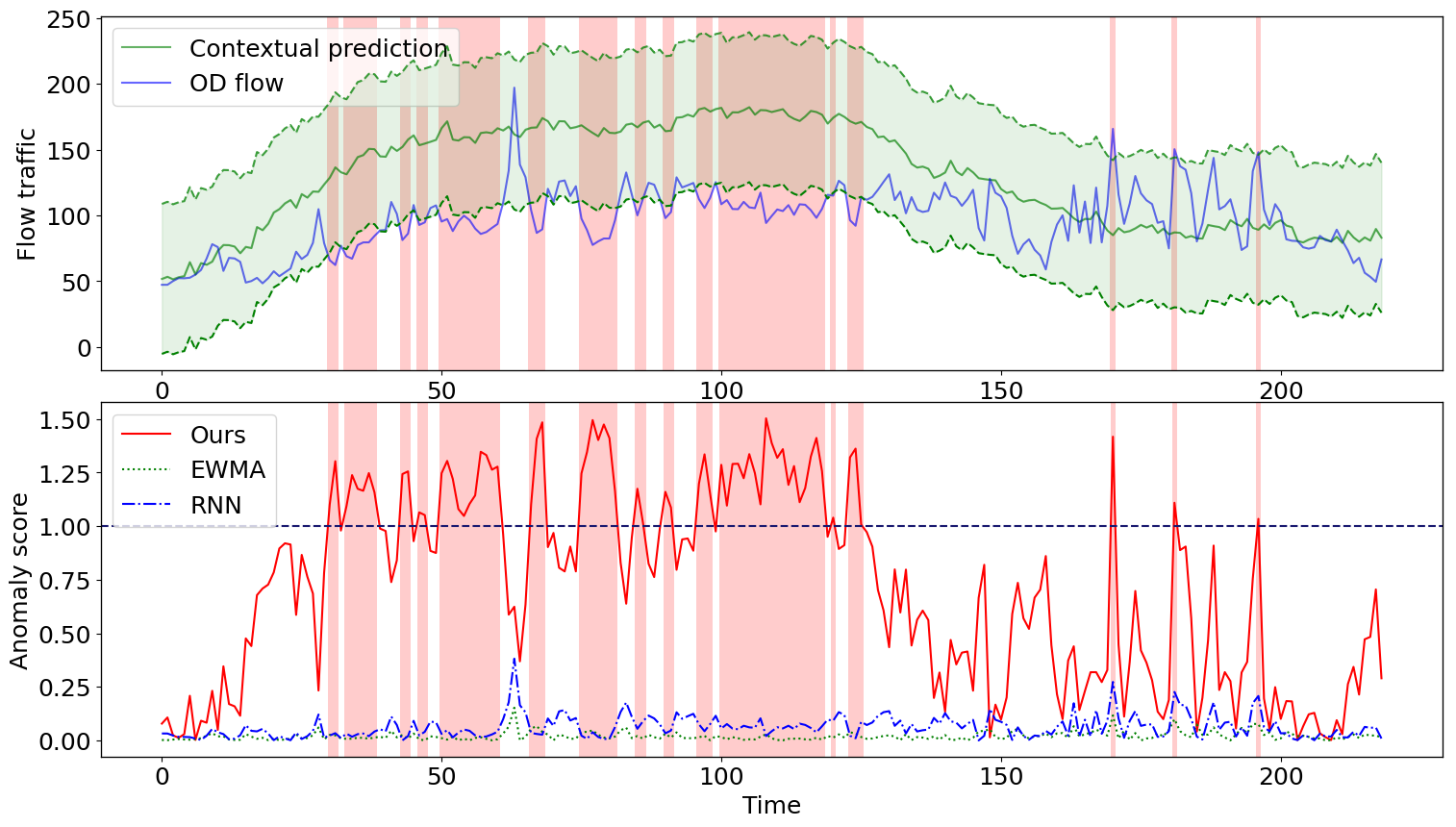}}
    \caption{Anomaly \#2 - Stable OD flow with shift on context flows.}
    \label{Anomaly 2 plots}
\end{figure}

Figures \ref{Anomaly 1 plots} and \ref{Anomaly 2 plots} respectively show two examples of both classes. The top plots in both figures represent the temporal evolution of the OD flow (blue), along with the contextual predictions produced by the GNN model (green). Confidence intervals represent the anomaly detection thresholds considered in our evaluation (based on the behaviour of the flow context). All values exceeding these intervals are marked as anomalous. Likewise, the bottom plots in both figures represent the anomaly scores produced by the different methods over time. Note that PCA-based methods do not produce anomaly scores at the OD flow level, instead they annotate network-wide anomalies on time bins. Hence, we exclude them from this analysis. Since the range of anomaly scores is different for the different methods, we normalize scores across methods for a fair comparison, so that 1.0 represents the detection threshold for all of them. Note that thresholds were carefully set to produce the top 600 anomalies on each method. 
In both examples we can observe that the anomaly scores generated by our solution increase considerably when the OD flow and the contextual predictions start to deviate. This is because our method is tailored to amplify the detection of contextual anomalies. Conversely, the anomaly scores produced by the baselines are moderate and quite far from their detection thresholds. This can be explained by the fact that they do not capture deviations with respect to the context.

\section{Conclusion}

In this paper we presented a novel approach for contextual network anomaly detection based on Graph Neural Networks. We evaluated our solution with real-world traffic measurements from the Abilene backbone network. 
By comparing the detections produced by our method to other four baselines, we show that our solution is quite complementary to more traditional methods. This is mainly because our solution is specifically designed to amplify contextual anomalies, while the other methods use different detection algorithms based on time series analysis (EWMA, RNN) and data point outlier detection (PCA). These results seem promising towards developing DL-based methods that can accurately detect contextual network anomalies, thus complementing existing non-contextual methods.

As future work, we plan to dig deeper into the properties of the anomalies generated by our solution, by making a validation with annotated real-world network data. This would allow us to evaluate the effectiveness of our method in terms of precision and recall, as well as to further tune it to produce a low false alarm rate. Likewise, we plan to explore the use of other DL-based methods that may also be suitable for this task, such as Transformer models.

\begin{acks}
This publication is part of the Spanish I+D+i project TRAINER-A (ref.~PID2020-118011GB-C21), funded by MCIN/AEI/10.130\-39/501100011033. This work is also partially funded by the Catalan Institution for Research and Advanced Studies (ICREA). This work was supported by the Spanish Ministry of Economic Affairs and Digital Transformation and the European Union – NextGeneration EU, in the framework of the Recovery Plan, Transformation and Resilience (PRTR) (Call UNICO I+D 5G 2021, refs. number TSI-063000-2021-3,  TSI-063000-2021-38, and TSI-063000-2021-52). It has also received funding from the European Union’s Horizon 2020 research and innovation program under grant agreement no. 101017109 “DAEMON”.
\end{acks}

\bibliographystyle{ACM-Reference-Format}
\balance
\bibliography{main}

\end{document}